\documentclass[letterpaper]{article} 
\usepackage{aaai25}  
\usepackage{times}  
\usepackage{helvet}  
\usepackage{courier}  
\usepackage[hyphens]{url}  
\usepackage{graphicx} 
\usepackage{natbib}  
\usepackage{caption} 
\usepackage{algorithm}
\usepackage{algorithmic}
\usepackage{amssymb}
\usepackage{amsmath}
\usepackage{multirow}
\usepackage{booktabs}
\usepackage[autostyle=true,french=guillemets]{csquotes}
\usepackage{enumitem}
\usepackage{amssymb} 
\usepackage{newfloat}
\usepackage{listings}
\DeclareCaptionStyle{ruled}{labelfont=normalfont,labelsep=colon,strut=off} 
\floatstyle{ruled}
\newfloat{listing}{tb}{lst}{}
\floatname{listing}{Listing}
\title{Video Anomaly Detection with Motion and Appearance Guided \\Patch Diffusion Model}
\author{
    Hang Zhou,
    Jiale Cai,
    Yuteng Ye,
    Yonghui Feng,
    Chenxing Gao,\\
    Junqing Yu,
    Zikai Song\thanks{indicates corresponding author.},
    Wei Yang 
}
\affiliations{
    Huazhong University of Science and Technology, Wuhan, China\\
    \{henrryzh, jaile\_cai, yuteng\_ye, yhfeng, cxg, yjqing, skyesong, weiyangcs\}@hust.edu.cn
}

\usepackage{bibentry}

\begin{document}

\maketitle

\begin{abstract}
A recent endeavor in one class of video anomaly detection is to leverage diffusion models and posit the task as a generation problem, where the diffusion model is trained to recover normal patterns exclusively, thus reporting abnormal patterns as outliers. Yet, existing attempts neglect the various formations of anomaly and predict normal samples at the feature level regardless that abnormal objects in surveillance videos are often relatively small.
To address this, a novel patch-based diffusion model is proposed, specifically engineered to capture fine-grained local information. We further observe that anomalies in videos manifest themselves as deviations in both appearance and motion. Therefore, we argue that a comprehensive solution must consider both of these aspects simultaneously to achieve accurate frame prediction. To address this, we introduce innovative motion and appearance conditions that are seamlessly integrated into our patch diffusion model. These conditions are designed to guide the model in generating coherent and contextually appropriate predictions for both semantic content and motion relations. Experimental results in four challenging video anomaly detection datasets empirically substantiate the efficacy of our proposed approach, demonstrating that it consistently outperforms most existing methods in detecting abnormal behaviors.
\end{abstract}

%

\section{Introduction}
Video anomaly detection (VAD) remains a challenging problem within the security monitoring domain~\cite{Lu, dmu, XD}. Due to the scarcity of anomalous data and restrictions on data collection, training predominantly employs videos depicting normal behavior. To address this imbalance, prevailing strategies have focused on frame prediction and reconstruction methods~\cite{Stacked-RNN, Zhong-et-al, FFP, compact}. During testing, anomalous frames are identified based on significant discrepancies between predicted and actual frames. In this paper, we adopt the frame prediction approach to generate future video frames~\cite{LMCMem}.
\begin{figure}[t]
    \centering
    \includegraphics[width=0.95\columnwidth,page=1]{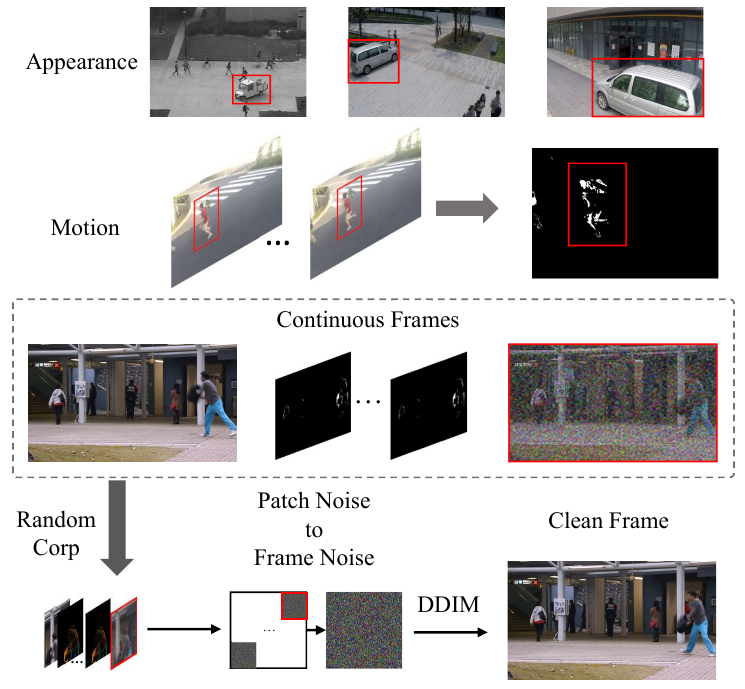} 
    \caption{We propose a patch-based diffusion model with a motion and appearance conditions framework for VAD. The conditional frames, which contain motion of temporal difference and appearance information, are cropped into patches and used to estimate noise. These noises are then combined and refined into the frame-level noise. The DDIM reverse step is then applied to recover the clean image.}
    \label{fig: teaser}
\end{figure}

The autoencoder (AE) model is the most widely used technique for VAD~\cite{MemAE, CDDA}. However, its remarkable generalization capability can inadvertently lead to the reconstruction of anomalous behavior. To mitigate this overgeneralization, several methods, such as MNAD~\cite{MNAD} and MPN~\cite{MPN}, introduce learnable and dynamic memory blocks to store various normal patterns. Nevertheless, the placement of the memory unit within the bottleneck of AE, along with the use of skip connections, still permits the reconstruction of abnormal patterns.
Alternative approaches, such as AMMC-Net~\cite{AMMC-Net}, aim to diminish the capacity of the model by integrating additional motion information. This method employs a two-stream memory augmentation network, merging motion and appearance prior knowledge to enhance multi-view features for regular events. However, its performance in anomaly detection is restricted due to the powerful autoencoder with U-Net architecture~\cite{unet}. An attempt has been made to use the diffusion model to generate video frames without the need for capacity reduction (unlike AE) ~\cite{DDPM, DDIM}. FPDM~\cite{fpdm} propose a feature reconstruction method with a two-stage diffusion model in the latent space. However, it is not able to capture the motion and appearance information of minor irregularities with coarse image features.

In this paper, we propose a novel Patch-based Diffusion Model with Motion and Appearance Conditions (MA-PDM), which is shown in Figure \ref{fig: teaser}. We decompose video frames into separate appearance and motion components. The initial frame encapsulates all visual aspects of appearance, thereby elevating the challenge of video frame prediction, whereas successive motion frames chronicle the temporal dynamics.
Consequently, our model distinctively exploits the intrinsic appearance and motion information present in videos as organic conditions, facilitating the generation of frames with enhanced precision and control.
To enhance content understanding, we introduce an appearance encoder equipped with a memory block that aids the diffusion model in comprehending appearance information. Drawing inspiration from the Condition Block of ControlNet~\cite{ControlNet}, we use a residual connection to integrate the appearance information with the noise estimation network. Moreover, we propose a prior motion strategy to guide diffusion models, utilizing temporal differences to gain a finer granularity of motion discrepancies.

In particular, as anomalous objects or behaviors in surveillance videos tend to be small and localized, a comprehensive image sequence analysis with a diffusion model may be suboptimal. To address this, we adopt patch-based techniques, which have proven effective in image anomaly detection to achieve precise localizations~\cite{PatchCore, csw}. Following the patch partition approach~\cite{PatchWDM}, we develop a patch-based diffusion model for precise detection and propose a fusion strategy for the conditions mentioned above.

In summary, our primary contributions are as follows:
\begin{itemize}
\item We innovate a patch-based condition diffusion framework designed to precisely capture local anomalous activity and ensure the normal area.
\item We introduce a motion and appearance conditioning strategy for the diffusion model to predict normal frames with controllable motion and appearance.
\item We propose a novel fusion method for both conditions, enabling the model to learn and differentiate normal appearance and motion patterns. In this context, the semantic information generated by the patch memory bank is strategically embedded to ensure a significant separation between normal and abnormal segments.
\end{itemize}

\section{Related Work}
\label{sec:Related Work}
\subsection{Video Anomaly Detection}
Due to data collection limitations, existing approaches learn normal patterns to distinguish abnormal behavior that violates normality~\cite{AnoPCN}. These methods can be categorized into two groups: frame-based and object-center-based methods. 

\textbf{Frame-based VAD.}
Frame-based methodologies commonly utilize an autoencoder (AE) structure to forecast upcoming frames by leveraging previous ones. For instance, a 3D-AE architecture is presented to capture local spatial and temporal receptive fields. AMC integrates AE with U-Net to produce future frames and optical flow, thereby improving network representation. Researchers have explored the remarkable generalization capabilities of AE models, enabling the accurate prediction of abnormal frames. To enhance the variety of normal patterns, MNAD introduces a more compact and sparse memory module. To optimize storage efficiency and improve relevant features, DLAN-AC introduces a dynamic local aggregation network with an adaptive cluster. While memory-based strategies effectively maintain normality, they are limited by the AE model's inclination to oversimplify. Recently, diffusion models have emerged with robust and adjustable generation capabilities. FPDM, a feature prediction approach based on a two-stage diffusion model, is suggested to reduce noise in future frame features~\cite{fpdm}. The feature space is coarse for accurately reconstructing future frames, making it less suitable for identifying anomalous regions. Combining motion and appearance details is considered advantageous for diffusion models in predicting normal frames.

\textbf{Object-centered VAD.} 
Object-oriented techniques initially identify the object of interest in the foreground, then apply the same prediction or reconstruction techniques, significantly enhancing the robustness of VAD. For example, recent works~\cite{VEC, BDPN} forecast patches related to video events based on detected objects, thus avoiding background interference. However, network performance is restricted by the single prediction or reconstruction task. To improve the embedding performance of the U-Net model, HF2VAD~\cite{hf2vad} combines prediction and reconstruction tasks to incorporate a high correlation between RGB and the optical flow of objects. Furthermore, HSNBM~\cite{HSNBM} argues that real-world objects are associated with the entire background and introduces a hierarchical scene normality-binding model to enhance frame prediction performance. 
In summary, patch-based methods mainly concentrate on detected objects, but can be time-consuming when there are many objects. We incorporate the image patch way into our diffusion model to focus on local areas and use sliding windows for the testing phase to reconstruct frames.

\begin{figure*}[t]
    \centering
    \includegraphics[width=0.9\textwidth,page=2]{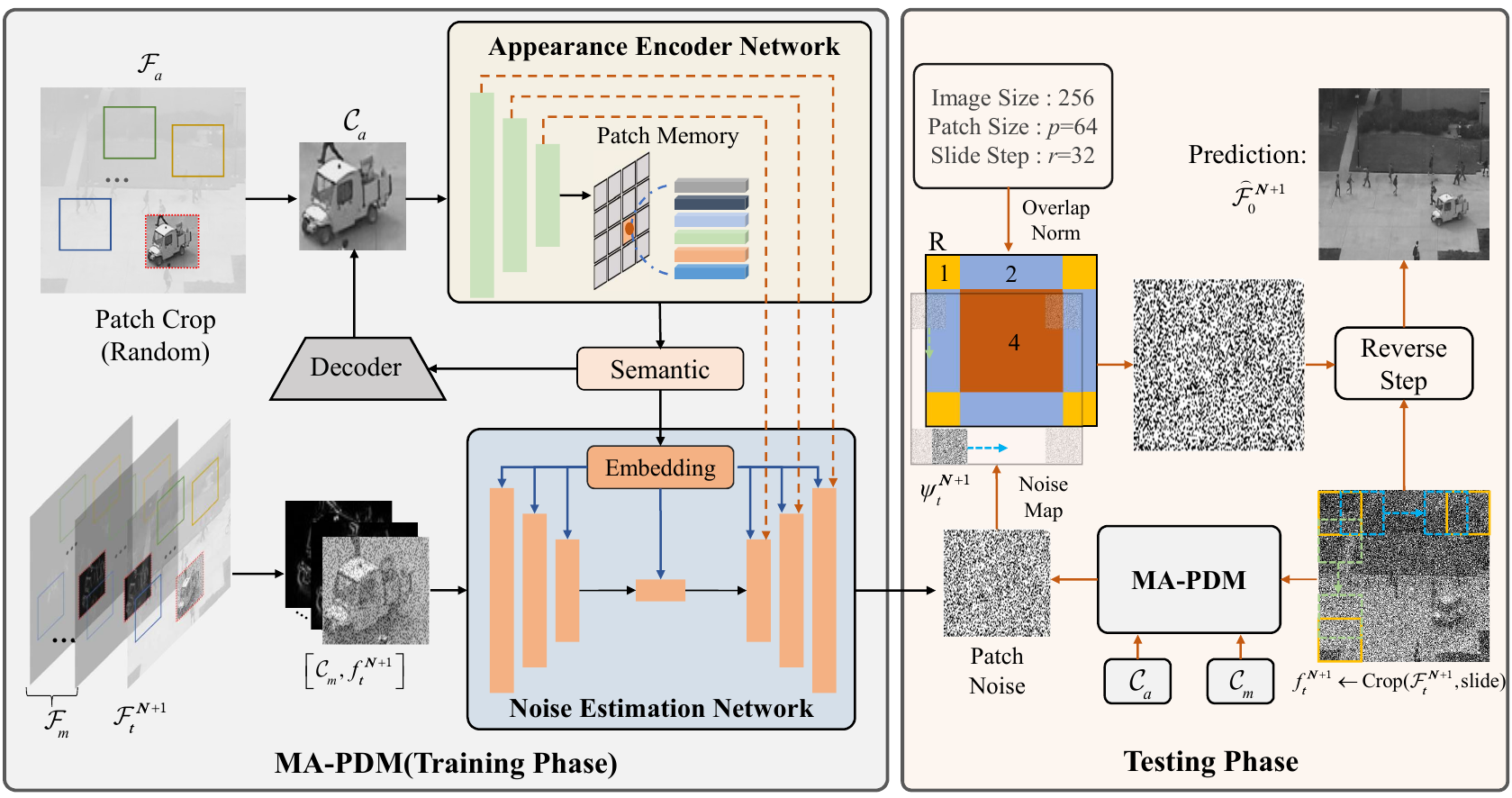} 
    \caption{Our MA-PDM comprises three components: a patch cropping module for creating patch conditions and noise, an appearance encoder for embedding and retaining the regular pattern, and a noise estimation network for forecasting the noise. During the training stage, the MA-PDM is trained to predict forward noise. During the inference stage, the MA-PDM anticipates the patch noise using conditions and then combines them in a reverse process. 
    } 
    \label{framework}
\end{figure*}

\subsection{Diffusion Model}
Recently, there has been a growing interest in diffusion models in the field of generative models~\cite{ye, DDIM, LDM}, which have demonstrated high effectiveness in image synthesis. These models are built on a Markov chain that gradually learns the data distribution by undergoing a forward diffusion process and subsequently learns the reverse diffusion process to reconstruct it. Conditional generative models based on diffusion have achieved impressive results in various tasks, such as synthesizing class-conditional data with the guidance of classifiers~\cite{conddm}, enhancing image resolution~\cite{SR3}, deblurring images~\cite{PatchWDM, repaint}, generating and editing images based on text descriptions~\cite{LDM, Imagen, diffedit, NullText}, and performing general image-to-image translation tasks~\cite{DAE, HDAE}. These investigations indicate the importance of conditions for the diffusion model to produce high-quality images. Moreover, diffusion models are also being explored for anomaly detection. In the medical field, \cite{MedDM} introduce the diffusion model for detecting medical anomalies with the guidance of classifiers. \cite{ADDM} propose a one-step diffusion denoising process during inference, achieving excellent performance in industry anomaly detection. Unlike detecting anomalies in images, video anomaly detection focuses on identifying irregular movements or objects. However, due to background noise and small targets, the diffusion model may lose its effectiveness in generating realistic outputs without well-designed conditions. In this study, we introduce a patch-based diffusion model with prior conditions, specifically motion and appearance cues, to address both aspects in videos.

\section{Method}
We introduce a new method for VAD, termed the Motion and Appearance Conditions Patch-based Diffusion Model (MA-PDM). Unlike previous methods that rely on an AE model for frame generation, our approach utilizes a diffusion model for forecasting future frames. Illustrated in Figure \ref{framework}, the MA-PDM comprises three key elements: patch cropping, appearance encoder, and noise estimation network. Our study is structured into training and testing stages. In the subsequent sections, we will initially discuss the problem setup, followed by an explanation of the three aforementioned modules, and conclude with an overview of the training and testing procedures.

\subsection{Problem Formation}
Following previous work, we regard the VAD task as a future frame prediction task.
In the training phase, we solely use normal video data and leverage the preceding frames $\{\mathcal{F}^{1},\dots,\mathcal{F}^{N}\}$ to forecast the subsequent frame $\mathcal{F}^{N+1}$. During the testing stage, we employ the diffusion model to recover the forthcoming frame with Gaussian noise and reconstruct them from patch size to their original resolution. Subsequently, we will elaborate on the specifics of our diffusion model with the prior constraints of videos.

\subsection{Diffusion Model with Motion and Appearance Conditions}
Diffusion models propose a noise estimation function $\epsilon_{\theta}(x_t,t)$ that takes a noisy image $x_t$ in a forward process and then predicts its noise. 
The model is trained by an MSE loss $||\epsilon_{\theta}(x_t,t)-\epsilon||_2^2$, where $\epsilon$ is the Gaussian noise added to the origin image $x_0$ to obtain $x_t$.
The same as these diffusion models, we define a Gaussian diffusion process at time step $t=1,\dots ,T$ and add the noise to our target frame $\mathcal{F}^{N+1}$ as: 
\begin{equation}
    \begin{aligned}
       q(\mathcal{F}^{N+1}_{t}|\mathcal{F}^{N+1}_{t-1}) = \mathcal{N}(\mathcal{F}^{N+1}_{t};\sqrt{1-{\beta}_t}\mathcal{F}^{N+1}_{t-1},{\beta_t} \textbf{I}),
    \end{aligned}
    \label{eq1}
\end{equation}
where $\beta_t \in(0,1)$ is the variance schedule of each time step. 
To acquire the data distribution $q(x_0)$ at time $t$, we can adapt the Eq.\ref{eq1}  as:
\begin{equation}
    \begin{aligned}
       q(\mathcal{F}^{N+1}_{t}|\mathcal{F}^{N+1}_{0}) = \mathcal{N}(\mathcal{F}^{N+1}_{t};\sqrt{{\alpha}_t}\mathcal{F}^{N+1}_{0},(1-{\alpha}_t) \textbf{I}),
    \end{aligned}
    \label{eq2}
\end{equation}
where ${\alpha}_t = \prod_{s=1}^t(1-\beta_s)$.
The noise step transforms our future frame at any noise step into a simple Gaussian distribution.

Conditions play a crucial role in the generation of images. The primary characteristics of video frames are motion and appearance, and we leverage both prior knowledge and diffusion conditions to steer the creation of normal patterns. Video frames serve as the fundamental appearance keyframes that we can choose, while there are two options to consider for motion information. We prioritize the temporal difference approach due to its ability to offer more intricate motion details under fixed surveillance perspectives and its quicker processing compared to optical flow. Consequently, we utilize the initial frame $\mathcal{F}_a = \mathcal{F}^1$ for appearance information and the temporal difference $\mathcal{F}_m$ between consecutive frames $\mathcal{F}=\{\mathcal{F}^1,\dots,\mathcal{F}^N\},N=6$, which can be mathematically represented as:
\begin{equation}
    \begin{aligned}
       \mathcal{F}_m=\mathcal{F}[:-1]-\mathcal{F}[1:].
    \end{aligned}
    \label{eq3}
\end{equation}

By incorporating both conditions, we can utilize a noise estimation network to infer the noise in the forward process. The loss function is specified as:
\begin{equation}
    \begin{aligned}
       \mathcal{L}_{simple}=\mathbb{E}_{t,f^{N+1}_0,\epsilon}[||\epsilon_{\theta}(F^{N+1}_t,t,\{\mathcal{F}_a,\mathcal{F}_m\})-\epsilon||^2],
    \end{aligned}
    \label{eq4}
\end{equation}
where $\epsilon_{\theta}(F^{N+1}_t,t,\{\mathcal{F}_a,\mathcal{F}_m\})$ is our noise estimation network with appearance $\mathcal{F}_a$ and motion $\mathcal{F}_m$ conditions. Details of the architecture will be introduced in Sec.\ref{arch}.

\subsection{Patch-based Diffusion Model}
The diffusion model is highly efficient in various scenarios, yet it necessitates a substantial amount of computational time for the iterative processing of a complete image. This becomes particularly challenging when working with large images, leading to significant delays in inference time. Moreover, the majority of surveillance images feature a background, posing a challenge for the diffusion model in discerning human objects across the entire image. To tackle this challenge, we introduce a patch-based diffusion model that assesses local patch noise to expedite the computation process while concentrating on more localized regions. In the training phase, we extract patch frames from all images and acquire the corresponding patches for different conditions and noisy images. To enhance the diversity and robustness of the patch approach, random boxes are generated to improve noise learning. This can be formulated as:
\begin{equation}
    \begin{aligned}
       \mathcal{C}_a &= \text{Crop}(\mathcal{F}_a,\text{random}),\\
       \mathcal{C}_m&= \text{Crop}(\mathcal{F}_m,\text{random}),\\
       f_t^{N+1} &= \text{Crop}(\mathcal{F}_t^{N+1},\text{random}).\\
    \end{aligned}
    \label{eq5}
\end{equation}

Hence, the forward process will also be carried out in a patch fashion, and the loss function is modified as:
\begin{equation}
    \begin{aligned}
       \mathcal{L}_{p}=\mathbb{E}_{t,f^{N+1}_0,\epsilon}[||\epsilon_{\theta}(f^{N+1}_t,t,\{\mathcal{C}_a,\mathcal{C}_m\})-\epsilon||^2].
    \end{aligned}
    \label{eq6}
\end{equation}

The training phase of our model predicts the noise of a local patch from the entire image by utilizing random cuts. Subsequently, we employ a sliding-window crop technique to generate mesh patches for the inference stage, which is discussed in Sec.\ref{infer}.

\subsection{Architecture}
\label{arch}
Figure \ref{framework} shows the architecture of our MA-PDM, which is composed of the appearance encoder and noise estimation network.
At the initial data processing stage, we combine the patch conditions of appearance $\mathcal{C}_a$ and motion $\mathcal{C}_m$. 
Following previous work on the image-to-image task~\cite{Medsegdiff, repaint}, we join the motion frames with the noisy frame $[\mathcal{C}_m,f_t^{T+1}]$ to form the input and then use the U-Net model to predict the adding noise.
The appearance patch $\mathcal{C}_a$ contains abundant semantics for the prediction task, and we use the same encoder part of the U-Net framework to extract the appearance and semantic information. 
Both information will be integrated into the U-Net as additional conditions. 

\textbf{Appearance Encoder Network.} As mentioned by DAE~\cite{DAE}, semantics play a crucial role in generating semantic content. By adopting a similar embedding approach and incorporating additional encoder features to handle hierarchical structures~\cite{altrack}, our appearance encoder becomes capable of capturing semantic embeddings and hierarchical attributes from the appearance context. This can be formally represented as:
\begin{equation}
    \begin{aligned}
       \relax [h_1,\dots,h_n]= \phi_e(\mathcal{C}_a),
    \end{aligned}
    \label{eq7}
\end{equation}
where $h_i$ represents the feature map of $i$-th layer of the encoder. 
$\phi_e$ is the appearance encoder network for embedding the input frame and has the same structure as the denoising U-Net.
In order to learn the normal semantics for only regular data, we utilize the learnable patch-memory bank $\mathcal{M}\in\mathbb{R}^{P\times N\times D}$ to preserve the normal semantics for different partial patches.  
The addressing operation is characterized as follows:
\begin{equation}
    \begin{aligned}
      \hat{\mathcal{C}}_{sem} = \text{CrossAtt}(W_q(h_n), W_k(\mathcal{M}_p), W_v(\mathcal{M}_p))
    \end{aligned}
    \label{eq8}
\end{equation}
where $h_n$ represents the output of the Appearance Encoder, $\mathcal{M}_p\in\mathbb{R}^{N\times D}$ denotes the $p$-th memory bank of $\mathcal{M}$, and $p$ corresponds to the patch location within the entire image. For instance, if the coordinates of the random patch $\mathcal{C}_a$ are $(58,58)$, then $p=\left\lceil \frac{58}{64} -\frac{1}{2} \right\rceil \times 4 + \left\lceil \frac{58}{64}-\frac{1}{2}\right\rceil=5$. Detail framework is shown in Supplementary Material. Depending on the addressing stage using the cross-attention module, the filtered semantic $\hat{\mathcal{C}}_{sem}$ is obtained to reconstruct the input patch $\mathcal{C}_a$ with a decoder network, and the reconstruction loss is applied to constrain the decoder result $\hat{\mathcal{C}_a}$:
\begin{equation}
    \begin{aligned}
       \mathcal{L}_{r}=\text{MSE}(\hat{\mathcal{C}_a},\mathcal{C}_a).
    \end{aligned}
    \label{mm_recloss}
\end{equation}


\textbf{Noise Estimation Network.}
The U-Net structure serves as the foundation for noise estimation. To incorporate motion and appearance conditions into the network, we initially combine the frame differences with the noise frame using $[\mathcal{C}_m,f_{t}^{N+1}]$ as input frames. Subsequently, the hierarchical feature maps $[h_1,\dots,h_n]$ are integrated into the U-Net network, detail framework is shown in Supplementary Material.
Filtered semantics $\mathcal{S} = \text{Pool}(\hat{\mathcal{C}}_{sem})$ are embedded in the entire network with the time-step embedding $t_{emb}$, the semantics are embedded in each ResBlock of U-Net by the following formulation:
\begin{equation}
    \begin{aligned}
        \mathcal{E}(h,t_{emb},\mathcal{S}) = (1+\phi_s(t_{emb}+\mathcal{S})) h+\phi_b(t_{emb}+\mathcal{S}),
    \end{aligned}
    \label{embed}
\end{equation}
where $\phi_s,\phi_b$ are the scale and bias parampters of Adaptive Layer Normalization.

In the phase of decoding part of the noise estimation network, we incorporate the hierarchical appearance feature of the appearance encoder network into the U-Net decoder, similar to ControlNet~\cite{ControlNet}. It is mathematically represented as $[h_d^N,h_e^N+h_e^A]$,
$h_d^N,h_e^N$ are the noise feature map derived from the encoder and decoder section of the noise estimation network. 
$h_e^A$ is the feature map $h_i$ sourced from the appearance encoder. 
$h_e^N+h_e^A$ is a dynamic skip connection to control appearance information.
Detail framework is shown in Supplementary Material.

\textbf{Loss function.}
At the training stage, we use the above two losses to learn our MA-PDM model:
\begin{equation}
    \begin{aligned}
        \mathcal{L} = \mathcal{L}_p + \lambda_1 \mathcal{L}_r,
    \end{aligned}
    \label{loss}
\end{equation}
\subsection{Inference Stage}
\label{infer}
During the inference phase, the reverse process of the diffusion model is used to extract random Gaussian noise under specific appearance and motion conditions. In contrast to the random style utilized during the training phase, images are generated from a standard partition. We use a sliding window to obtain various patches $f_{b}^{N+1}$ from the complete Gaussian noise $\mathcal{F}_{T}^{N+1}$, for example, the patch size is $p=64$, the sliding step is $s=r,r<p$, the total image size is $H=W=256$, and the position box $b$ is $(x_1,y_1,x_2,y_2)$.
There is an overlap in the patches due to the sampling strategy, and we compute the overlap pixel rate $R_i = n, R\in\mathbb{R}^{H\times W}$ as~\cite{PatchWDM}. Here, $R$ is the normalization matrix of the current sampling strategy, and $n$ is the number of pixels of overlap.
Using these patch samples $f_{t;b}^{N+1} \in \mathbb{R}^{c \times p \times p}$ generated in the position box $b$, we predict the noise $\epsilon_b$ at step $t$ using our MA-PDM network:
\begin{equation}
    \begin{aligned}
    \epsilon_b = \epsilon_{\theta}(f^{N+1}_{t;b},t,\{\mathcal{C}_a,\mathcal{C}_m\}).
    \end{aligned}
    \label{eq9}
\end{equation}
We assemble the patches $\{\epsilon_b\}$ to build the complete image noise $\psi _{t}^{N+1}$ using the subsequent merging technique:
\begin{equation}
    \begin{aligned}
    &\psi _{t}^{N+1}[b] = \epsilon_b,\quad \psi _{t}^{N+1}  = \frac{\psi _{t}^{N+1}}{R}.
    \end{aligned}
    \label{eq10}
\end{equation}
With the matrix of $R$, we can always maintain the stable amplitude of the noise.

The implicit sampling \cite{DDIM}, which exploits a generalized non-Markovian forward process formulation, is applied to our diffusion model to accelerate the sampling speed.
The reverse process is defined as:
\begin{equation}
    \begin{aligned}
    &q(\mathcal{F}_{t-1}^{N+1} \vert \mathcal{F}_{t}^{N+1},\{\mathcal{F}_a,\mathcal{F}_m\}) \\
    &= \mathcal{N}(\mathcal{F}_{t-1}^{N+1};\mu_{\theta}(\mathcal{F}_t^{N+1},t,\{\mathcal{F}_a,\mathcal{F}_m\});\beta_t \textbf{I}),
    \end{aligned}
    \label{eq11}
\end{equation}
where $\mu_{\theta}(\mathcal{F}_t^{N+1},t,\{\mathcal{F}_a,\mathcal{F}_m\})$ indicates the estimate of denoising transition mean in the forward process.
Hence, a sample step can be generated from the DDIM reverse process on both conditions by:
\begin{equation}
    \begin{aligned}
    \mathcal{F}_{t-1}^{N+1} 
    =& \sqrt{\alpha_{t-1}} \left( \frac{\mathcal{F}_{t}^{N+1}-\sqrt{1-\alpha_t} \cdot \psi _{t}^{N+1}}{\sqrt{\alpha_t}} \right) \\
    &+ \sqrt{1-\alpha_{t-1}} \cdot  \psi _{t}^{N+1},
    \end{aligned}
    \label{eq12}
\end{equation}

Finally, we can obtain the clean image $\hat{I}_{N+1}=\hat{\mathcal{F}}_0^{N+1}$ after all the denoising steps. 
Similarly, we can also merge the appearance conditions of the reconstruction $\hat{\mathcal{C}}_a$ to get $\hat{I}_1$ according to the above method.
We compute the anomaly score using the MSE function between the predicted frame ${\hat{I}_{N+1},\hat{I}_{1}}$ and its ground truth $I_{N+1},I_1$ as follows:
\begin{equation}
    \begin{aligned}
        \text{Score} = \text{MSE}(I_{N+1},\hat{I}_{N+1}) + \alpha \text{MSE}(I_{1},\hat{I}_1) 
    \end{aligned}
    \label{psnr}
\end{equation}


\section{Experiments}
\subsection{Experimental Setup}

\textbf{Datasets.}
We assess the effectiveness of our approach using four standard datasets prevalent in the Visual Anomaly Detection community. The datasets we employ include:
\begin{itemize}
    \item Ped2. It comprises 16 training videos and 12 test videos captured in static environments. 
    \item Avenue. It consists of 16 training videos and 21 testing videos, with a total of 47 abnormal events, including throwing a bag, approaching or moving away from the camera, and running on pavements.
    \item ShanghaiTech Campus. It consists of 330 training videos and 107 testing videos, which cover a variety of scenarios. It also contains 130 abnormal events, including brawls, robberies, fights, and other unusual activities.
    \item UBnormal. It composes of various virtual scenes created using the Cinema4D software with 29 scenes. We utilize the standard training and testing sets provided to assess the performance of our method for one-class VAD.
\end{itemize}

\textbf{Evaluation Metric.}
In order to evaluate the effectiveness of our approach, we employ the area under the curve (AUC) of the Receiver Operating Characteristic (ROC) curve at the frame-level, a commonly adopted assessment measure in the field of VAD.

\textbf{Training Details.}
Each frame is resize to the image size of $256\times 256$. The seventh frame is predicted using the previous six frames. The patch memory banks' capacity for all datasets is established at $16\times64\times256$. In the training phase, patches are randomly selected and extracted from complete images, with each patch measuring $64\times64$ for all datasets. The Adam optimizer is used to train our MA-PDM, with a learning rate set at $0.0002$. The diffusion process settings are as follows: $\beta_1=0.0001$, $\beta_2=0.02$, $T=1000$, and the linear schedule is implemented in the same way as in DDIM. The weights of the loss function are $\lambda= 0.1$. The training epochs are set to 1000 for Ped2, 300 for Avenue, 30 for Shanghai and 40 for UB. The batch size in all three datasets is 16. During the testing phase, we employ a sliding window approach with a stride of $64$. The number of reverse steps in DDIM is configured to $5$. The values for $\alpha$ are $(0,0.2,0.3,1)$ across four datasets.


\begin{table}[t]
\begin{center}
    \caption{
     Comparison of SOTA methods on four video anomaly datasets. The average AUC score ($\%$) is used as the measurement metric. Here, $\text{R.}$ and $\text{P.}$ denote reconstruction- and prediction-based methods, $*$ indicates our reproduction. The top two scores are highlighted with bold and underline.
    }
    \label{table:result_sota}
    \resizebox{0.99\columnwidth}{!}{
        \begin{tabular}{c|cccccc}
                \toprule
                                    & Method        & Venue                    & Ped2          & Avenue        & Shanghai      & UB \\
                \midrule
                \multirow{9}{*}{R.} & Conv-AE     & CVPR16       & 90.0          & 70.2          & 60.9         & - \\
                                    & ConvLSTM-AE  & ICME17  & 88.1          & 77.0          & -          & - \\
                                    & Stacked RNN & ICCV17   & 92.2          & 81.7          & 68.0          & -\\
                                    & AMC    & ICCV19                & 96.2          & 86.9          & -           & -\\
                                    & MemAE   & ICCV19             & 94.1          & 83.3          & 71.2          & -\\
                                    & CDDA   & ECCV20               & 96.5          & 86.0          & 73.3          & -\\
                                    & MNAD   & CVPR20               & 90.2          & 82.8          & 69.8          & -\\
                                    & Zhong et al. & PR22  & 97.7          & 88.9          & 70.7          & -\\ 
                                    & FPDM      & ICCV23            &  -            & \underline{90.1}          & 78.6          & \underline{62.7}\\ \hline
                \multirow{7}{*}{P.} & FFP      & CVPR18              & 95.4          & 84.9          & 72.8          & -\\
                                    & AnoPCN    & ACMMM19          & 96.8          & 86.2          & 73.6         & - \\
                                    & MNAD       & CVPR20           & 97.0          & 88.5          & 70.5         & $55.3^*$ \\
                                    & ROADMAP     & TNNLS21       & 96.3          & 88.3          & 76.6         & - \\
                                    & MPN       & CVPR21             & 96.9          & 89.5          & 73.8          & -\\
                                    & AMMC-Net  & AAAI21        & 96.9          & 86.6          & 73.7         & - \\
                                    & LGN-Net   & Arxiv23             & 97.1          & 89.3          & 73.0         & $57.4^*$ \\
                                    & DLAN-AC   & ECCV22         & 97.6          & 89.9         & 74.7          & -\\
                                    & USTN-DSC  & CVPR23        & \underline{98.1}          & 89.9          & 73.8         & - \\ 
                                    & AED-MAE    & CVPR24      & 95.4          & \textbf{91.3}         & \underline{79.1}         & 58.5 \\  \cline{2-7} 
                                    & \textbf{Ours}       & -               & \textbf{98.6} &\textbf{91.3} &\textbf{79.2}  & \textbf{63.4}\\ 
                \bottomrule
                \end{tabular}
        }
\end{center}
\end{table}

\subsection{Comparison with state-of-the-art}

In Table \ref{table:result_sota}, we present a comprehensive comparison of our innovative frame prediction-based approach compared to current best-performing methods in the domains of reconstruction and prediction on four benchmark datasets. Our evaluation includes various memory-based methods, particularly MPN~\cite{MPN}, LGN-Net~\cite{lgn}, and DLAN-AC~\cite{DLAN-AC}, known for their advanced anomaly detection capabilities. The results, as shown in the table, indicate that our approach outperforms all other methods assessed. Meanwhile, our MA-PDM model establishes a new standard by improving SOTA performance metrics by 1. 2\%, 0. 5\%, and 0. 7\% in the Avenue, Shanghai, and UB datasets, respectively, compared to the diffusion method: FPDM~\cite{fpdm}. We can also achieve performance that is close to or even better than AED-MAE~\cite{aed-mae} without any synthetic anomalies.

\begin{table}[t]
    \caption{
        Comparison of each module's performance on four video anomaly datasets.
        PD indicates our patch-based diffusion model, A-C indicates the appearance condition, and M-C indicates the motion condition.
        }
    \label{table:moudle}
    \begin{center}
        \resizebox{0.95\columnwidth}{!}{
            \begin{tabular}{ccc|cccc}
                \toprule
                PD              & A-C          & M-C        & Ped2          & Avenue        & Shanghai  & UB\\
                \midrule
                \checkmark      &              &            & 96.8          & 89.2          & 73.9     & 56.3 \\
                                & \checkmark   &            & 95.8          & 86.2          & 72.4     & 55.3 \\
                                & \checkmark   & \checkmark & 96.7    & 88.8   & 76.7    & 57.8\\
                \checkmark      & \checkmark   &            & 97.4    & 89.9    & 77.1   & 58.4  \\
                \checkmark      & \checkmark   & \checkmark & 98.6    & 91.3    & 79.2   & 63.4  \\ 
                \bottomrule
            \end{tabular}
        }
    \end{center}
\end{table}

\subsection{Ablation Study}

\textbf{Effectiveness of proposed modules.}
We present the results of five combinations of three modules to examine their respective influences, as detailed in Table \ref{table:moudle}. 
The Patch Diffusion model is efficient from the results of Case 1.
It illustrates that we have replicated the MNAD in the A-C scenario for Case 2, which only uses the six subsequent frames to predict the seventh frame using the memory block. 
As depicted in Case 3, a minor enhancement is observed when the motion condition is incorporated with MNAD.
Specifically, in Case 4, when integrated with a patch-based diffusion model, it enhances the performance on four datasets by $1.6\%$, $3.7\%$, $4.7\%$, and $3.1\%$ compared to Case 2 of MNAD that only uses the appearance condition.
This implies that the patch-based diffusion model is successful in handling irregular frames and achieving exceptional results.
The final case, utilizing our comprehensive model, shows the effectiveness of motion, allowing us to reach peak performance in detecting anomalies in videos.

\begin{table}[t]
    \caption{
        Comparison of the results of different patch sizes under two distinct cropping configurations.
        }
    \label{table:pathsize}
    \begin{center}
    \resizebox{0.95\columnwidth}{!}{
        \begin{tabular}{c|cccc|ccc}
        \toprule
        \multirow{2}{*}{Dataset} & \multicolumn{4}{c|}{Random Crop}                             &\multicolumn{3}{c}{Uniform Crop} \\ \cline{2-8} 
                                    & 32        & 64               & 128       & 256         & 32     & 64           & 128 \\ 
        \midrule
        Ped2                        & \textbf{98.6}      & \textbf{98.6}    & 97.2      & 96.8        & 97.7   & 97.3         & 96.2  \\
        Avenue                      & 90.2      & \textbf{91.3}    & 89.9      & 88.7        & 88.7   & 89.4         & 89.1   \\ 
        Shanghai                    & 78.7      & \textbf{79.2}    & 78.9      & 74.8        & 73.3   & 77.6         & 76.4   \\
        \bottomrule
        \end{tabular}
        }
    \end{center}
\end{table}

\begin{figure*}[t]
    \centering
    \includegraphics[width=0.8\textwidth,page=3]{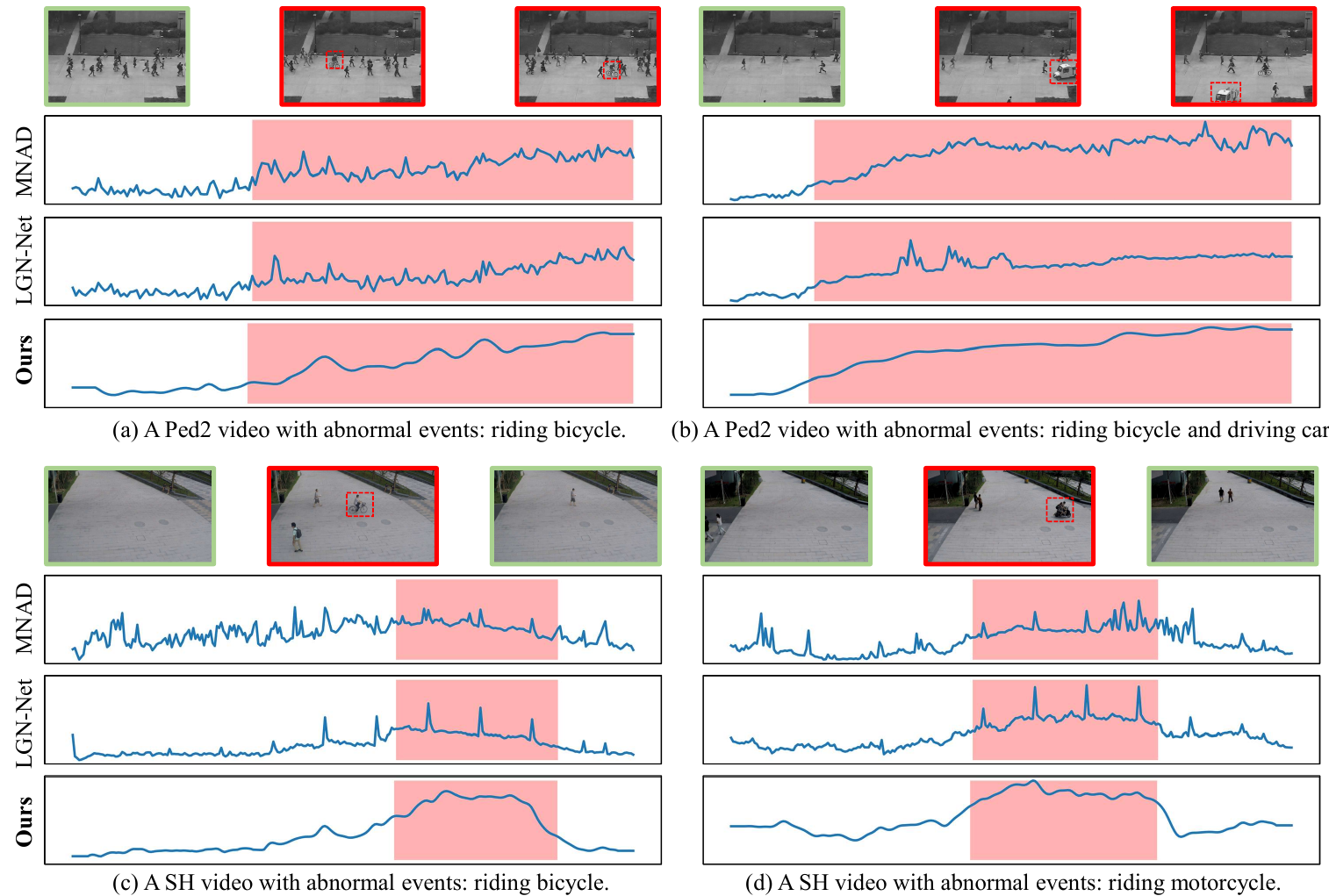} 
    \caption{
    Four examples of anomaly detection comparison on Ped2 and ShanghaiTech datasets.
    }
    \label{fig3}
\end{figure*}

\textbf{Effectiveness of patch-base diffusion model.}
We conduct an experiment to determine the optimal patch size for our patch-based diffusion model on three datasets. Initially, we evaluated four distinct patch dimensions $(32,64,128,256)$ using a random cropping approach. The results of the AUC score are shown in Table \ref{table:pathsize}. Our technique achieves the highest results on four datasets with a patch size of 64. Intriguingly, in the Ped2 and Avenue datasets, pedestrians are small and the scenes are singular, making the size of 32 more efficient for the Ped2 dataset. With an increased patch size of the complex dataset, the model will discern more data. The comprehensive image-based diffusion model, with a patch size of 256, yields a lower performance compared to the other patch-based diffusion models. 
The results of the uniform cropping method are then displayed in Table \ref{table:pathsize}. This suggests that a patch size of 32 is appropriate for the Ped2, while a size of 64 is suitable for the other dataset. The random cropping method is shown to be more efficient than the uniform one, as it increases and enlarges the dataset.

\textbf{Effectiveness of patch memory bank model.}
To explore the impact of memory block with patch style, we report the results of three types of memory variants in Table~\ref{tab: memory}.
Case 1 of \enquote{W/O} indicates the results obtained without the integration of any memory blocks. It achieves the approximate level compared with USTN-DSC~\cite{USTN-DSC}.
\enquote{G-M} represent the global memory without chunking, it has a larger memory size than our patch-based \enquote{P-M} method.
From the Case-2 and Case-3 results, it indicates the patch memory bank is more efficent for our patch diffusion model.

\textbf{Effectiveness of motion choice.}

We conduct an experiment to compare the performance of optical flow (OF) and temporal difference (TD). Table \ref{tab: motion comparison} displays the results of our experiment. We employ RAFT~\cite{raft} for the rapid and effective extraction of optical flows. Our TD method outperforms the OF technique in terms of results. In a comparison of the frames per second (FPS) for both techniques, our TD method proves to be quicker than the flow approach.  Evidently, the OF method requires additional time to extract features and is vulnerable to OF models.
\begin{table}[!t]
    \centering
    \caption{Comparison of the results of memory module.}
    \label{tab: memory}
    \resizebox{0.75\columnwidth}{!}{
        \begin{tabular}{c|ccc}
            \toprule
            Memory Variants  & Ped2              & Avenue            & Shanghai \\
            \midrule
            W/O      & 98.0                 & 89.9              & 77.9 \\
            G-M      & 98.6             & 90.2        & 78.7 \\
            P-M      & 98.6             & 91.3        & 79.2 \\
            \bottomrule
        \end{tabular} 
        }
\end{table}

\begin{table}[!t]
    \centering
    \caption{Comparison of the results of temporal difference (TD) and optical flow (OF) conditions.}
    \label{tab: motion comparison}
    \resizebox{0.75\columnwidth}{!}{
        \begin{tabular}{c|ccc}
            \toprule
            Motion  & Ped2              & Avenue            & FPS \\
            \midrule
            OF      & 93.7              & 88.7              & 25 \\
            TD      & 98.6(+4.9)        & 91.3(+2.6)        & 45(+20) \\
            \bottomrule
        \end{tabular} 
        }
\end{table}

\subsection{Qualitative Results}

Additionally, the anomaly curves of four test videos are shown in Figure \ref{fig3}, which offers a comparison between MNAD, LGN-Net and our MA-PDM. The curves represent the anomaly scores for each video frame in sequence, facilitating the comparison of the effectiveness of different methods. It's evident that our MA-PDM outperforms both MNAD and LGN-Net in abnormal sections, yielding higher and more stable anomaly scores. Furthermore, in normal periods, MA-PDM surpasses MNAD and LGN-Net, with our normal scores being more consistent than those of LGN-Net, as shown in Figure \ref{fig3}(a) and Figure \ref{fig3}(c). 
More visualizations are shown in Supplementary Material.

\section{Conclusion}
In this paper, we propose a Patch Diffusion Model with Motion and Appearance Guidance for VAD.
To capture fine-grained local information, we propose a novel patch-based diffusion model specifically engineered.
In addition, we introduce innovative appearance and motion conditions that are seamlessly integrated into our patch diffusion model.
Numerous experimental results on four benchmarks demonstrate that our model performs better than previous state-of-the-art methods.
Nevertheless, there is considerable room for improvement, particularly in the area of real-time speed for inference.
Our future work involves exploring faster frame prediction networks with the MA-PDM model and incorporating richer conditions, such as text prompts.

\section{Acknowledgment}
This work is supported by the National Key Research and Development Program of China (No.2020YBF2901202), National Natural Science Foundation of China (NSFC No. 62272184 and No. 62402189), the China Postdoctoral Science Foundation under Grant Number GZC20230894, the China Postdoctoral Science Foundation (Certificate Number: 2024M751012), and the Postdoctor Project of Hubei Province under Grant Number 2024HBBHCXB014. The computation is completed in the HPC Platform of Huazhong University of Science and Technology.

\bibliography{aaai25}

\end{document}